\documentclass[letterpaper]{article} 
\usepackage{aaai25}  
\usepackage{times}  
\usepackage{helvet}  
\usepackage{courier}  
\usepackage[hyphens]{url}  
\usepackage{graphicx} 
\usepackage{natbib}  
\usepackage{caption} 
\usepackage{algorithm}

\usepackage{newfloat}
\usepackage{listings}
\DeclareCaptionStyle{ruled}{labelfont=normalfont,labelsep=colon,strut=off} 
\floatstyle{ruled}
\newfloat{listing}{tb}{lst}{}
\floatname{listing}{Listing}
\usepackage{algpseudocode,amsmath,amssymb,multirow,subcaption,xcolor,tikz}
\usetikzlibrary{positioning,arrows.meta}

\usepackage[textsize=small,textwidth=3cm]{todonotes}
\definecolor{kentuckyblue}{RGB}{0, 93, 170}
\definecolor{union_garnet}{RGB}{134, 38, 51}

\newcommand{\x}{\mathbf{x}}
\newcommand{\X}{\mathbf{X}}

\newcommand{\rqone}{RQ1: How does removing label bias with ITE \textit{in the testing data} influence estimates of accuracy and fairness of classifiers trained on audit study data?}
\newcommand{\rqtwo}{RQ2: How does removing label bias with ITE \textit{in the training data} influence the accuracy and fairness of classifiers trained on audit study data?} 
\newcommand{\rqthree}{RQ3: How does the magnitude of discrimination in the human audit data influence the results?}
\newcommand{\rqfour}{RQ4: If audit study data is unavailable, how does selection bias affect the accuracy and fairness of resulting classifiers?}

\nocopyright
\title{The Illusion of Fairness: Auditing Fairness Interventions with Audit Studies}
\author {
    Disa Sariola,
    Patrick Button,
    Aron Culotta,
    Nicholas Mattei
}
\affiliations {
    Tulane University\\ New Orleans, LA, USA \\
    dsariola@tulane.edu, 
    pbutton@tulane.edu, 
    aronwc@tulane.com,
    nsmattei@tulane.edu
}

\begin{document}

\maketitle

\begin{abstract}

    Artificial intelligence systems, especially those using machine learning, are being deployed in domains from hiring to loan issuance in order to automate these complex decisions. Judging both the effectiveness and fairness of these AI systems, and their human decision making counterpart, is a complex and important topic studied across both computational and social sciences.
    Within machine learning, a common way to address bias in downstream classifiers is to resample the training data to offset disparities. For example, if hiring rates vary by some protected class, then one may equalize the rate within the training set to alleviate bias in the resulting classifier. While simple and seemingly effective, these methods have typically only been evaluated using data obtained through convenience samples, introducing selection bias and label bias into metrics.
    %
    %
    Within the social sciences, psychology, public health, and medicine, audit studies, in which fictitious ``testers'' (e.g., resumes, emails, patient actors) are sent to subjects (e.g., job openings, businesses, doctors) in randomized control trials, provide high quality data that support rigorous estimates of discrimination.  In this paper, we investigate how data from audit studies can be used to improve our ability to both train and evaluate automated hiring algorithms.
    %
    %
    %
    %
    %
    %
    We find that such data reveals cases where the common fairness intervention method of equalizing base rates across classes appears to achieve parity using traditional measures, but in fact has roughly 10\% disparity when measured appropriately. We additionally introduce interventions based on individual treatment effect estimation methods that further reduce algorithmic discrimination using this data. 
\end{abstract}

\section{Introduction}

A key assumption in most work on algorithmic discrimination is that the human provided class labels are improperly influenced, directly or indirectly, by a protected or minoritized characteristic like age, race, gender, or sexual orientation. Despite this assumption, most work evaluates machine learning methods on the very same flawed data, leading to biased estimates of both fairness and accuracy. While a truly unbiased annotation is infeasible for most tasks of interest (e.g., recidivism prediction, hiring, loans), in this paper we consider how this issue can be addressed using data from human audit studies\footnote{Audit studies are also known as audit field experiments, correspondence studies, resume studies/experiments, simulated patient studies, and vignettes -- henceforth ``audit studies'' for short.}

In an audit study, ``testers'' (e.g., resumes, inquiries, simulated patients) are randomly assigned covariates and protected characteristics and given to humans for assessment (e.g., to potentially offer an interview, response, or diagnosis). Due to this randomization, one can rigorously estimate the amount of discrimination in a real-world decision-making system \cite{gaddis2018introduction}. In fact, in an audit study, one would expect there to be no difference in the outcome (class) variable on average across the protected groups in the setting of no discrimination.

As our domain of study we analyze data from a real-world audit study of hiring that was collected to understand age discrimination in applicant callback decisions. Specifically, 40,208 resumes were sent to 13,371 job openings across 11 cities in the United States \cite{OlderWorkersFieldExperiment}, and information was recorded on which resumes received a callback from the company. The resumes were constructed systematically in order to isolate the effect of age on callback, controlling for other factors. 

Such a large and rigorously collected sample of human decisions serves as a useful testbed to investigate the behavior of machine learning systems trained to automate hiring decisions. In this paper, we study how such data may be used to improve both the training and evaluation of classifiers used in hiring and beyond. First, we explore how audit study data, unlike typical machine learning training data, enables robust estimates of the amount of \textit{label bias} in the data --- essentially, the quantity of human decisions that should be changed to eliminate discrimination. Second, we propose an approach based on individual treatment effect (ITE) estimation to quantify the likelihood that each individual record has been subject to discrimination (i.e. has a biased label). Based on these estimates, we then propose an algorithm to generate de-biased versions of the original data. We find that this approach allows us both to better estimate the true accuracy and fairness of classifiers, as well to improve the quality of classifiers trained on such data. 

Our empirical analysis focuses on the following four questions:
\begin{itemize}
    \item \textbf{\rqone} We find that traditional approaches can create an illusion of fairness, in which methods that appear fair when evaluated using standard approaches are in fact shown to exhibit significant discrimination once label bias is reduced.
    \item \textbf{\rqtwo} We find that training on debiased data reduces measures of disparity by up to 60\% as compared to traditional pre-processing approaches that equalize the base rate of callbacks across protected attributes.
     \item \textbf{\rqthree} When we sample data in such a way as to double the amount of human discrimination in the audit data, we find a commensurate increase in the inaccuracies of traditional fairness metrics. Our proposed ITE adjustment appears more robust, though classification accuracy does degrade in this setting.
    \item \textbf{\rqfour} When we reintroduce selection bias into the audit dataset, we find that illusion of fairness can become more extreme, with measures of fairness diverging not only in magnitude but also sign --- e.g., a classifier that appears discriminatory against younger applicants is in fact discriminatory against older applicants.
\end{itemize}

In the remainder of this paper, we first review related work in computational and social science (\S\ref{sec:related}), then introduce the audit study data and the machine learning task (\S\ref{sec:data}). We then illustrate the concept of label bias and how it can create an illusion of fairness (\S\ref{sec:illusion}), followed by our proposed methodology to address this problem using individual treatment effect estimates (\S\ref{sec:ite}). We then describe our experimental methodology (\S\ref{sec:experiments}) and empirical results (\S\ref{sec:results}), discuss the broader implications and limitations of this study (\S\ref{sec:discussion}), and conclude with future directions (\S\ref{sec:conclusion}).

\section{Related Work}
\label{sec:related}

In this section we provide an overview of the literature from the social sciences on audit studies and the use of these techniques in the fair machine learning literature.

\subsection{Audit Studies in the Social Sciences}

In the social sciences, psychology, public health, and medicine, audit studies generally refer to a specific type of field experiment in which a researcher randomizes characteristics about real, or hypothetical, individuals and sends them out into the field to test the effect of those characteristics on some outcome \cite{gaddis2018introduction}. In an audit study, ``testers'' (e.g., resumes, scenarios, emails) are randomly given the same characteristics (covariates), on average, except for those being tested for discrimination or differential treatment. A correspondence study is a type of audit study, where the audit is done through correspondence with subjects (e.g., emails, messaging, job applications) rather than in-person, such as with actors or simulated patients \cite{Collins2021}. Vignettes are another type of audit study that provides scenarios for subjects to consider (e.g., score applications, diagnose) \cite{Evans2015,Steiner2016}. While sometimes these are not ``real world'' situations, vignettes do allow for measuring discrimination or decision making in numerous cases that are not possible to study using actors or correspondence.

These approaches have become popular to apply to auditing \emph{algorithms} as well \cite{bandy2021algoaudit}. Algorithmic auditing is the application of social science auditing to a specific, identifiable system using both ``statistics and stories'' of given data in order to implement decision-making tools \cite{Vecchione_2021}. While the goals, methods, and outcomes of an algorithmic audit vary by context, the main emphasis is generally validation and verification of certain processes. These goals can vary: they might be with regard to transparency, or motivations of concerns with discrimination, or more traditional evaluations of performance in terms of accuracy, precision, and recall. Hence, the term auditing itself can refer to a multitude of different processes and techniques.  

The main difficulty when it comes to transitioning between social and algorithmic auditing comes from the challenges of translating concrete social criterion to a socio-technical implementation \cite{Vecchione_2021}. Some see automated machine learning and/or screening methods as inherently fair due to reflecting the underlying representations of the data. However, an audit cannot ever be truly fair in every step of the way, as the cumulative effect of small discriminatory actions at each step can add up to larger problems \cite{schumann2020we}. This becomes even more apparent when tight experimental controls and rigor become less transparent in the process due to the obfuscation caused by the inherent nature of computer algorithms. This does not mean that they are always worse, as there are some cases where an algorithmic audit is more impartial, as they are domain specific and concentrate on individual systems.

As noted, automated machine representations reflect the underlying data, and thus often have built-in bias stemming from said data.  However, bias can originate from multiple directions -- the model itself, the metrics used, or the data \cite{hutchinson201950}. A central goal of “fair machine learning” systems is to prevent such harm across groups \cite{10.1145/3514094.3534147}. This means that the result for subgroups should be comparable -- people of similar circumstances should receive similar outcomes independent of protected characteristics like gender, age, race or disability. See \citet{annurev:/content/journals/10.1146/annurev-statistics-042720-125902} for an overview of the broader impacts and definitions of what constitutes algorithmic fairness.


\subsection{Auditing Hiring}

The hiring process generally consists of three stages~\cite{2005116}. It begins with job planning, an analysis of the present and future needs for personnel and their competence for different tasks. Next, the employer makes a decision where and how to disseminate information about the opening. This includes the choice of venue, whether in person, physical fliers, or on online job-seeking sites, depending on the experience and type of personnel needed. In the third stage, the employer makes the decision on which of the applicants are considered for the opening, and which ones to hire. Our study focuses on the applicant selection process for callbacks.

The modern job market is highly competitive, with employers selecting from job-seekers, rather than applicants selecting from multiple offers \cite{RePEc:tpr:restat:v:67:y:1985:i:1:p:43-52}. According to a study based Swedish National Labor Market Board data starting in 1995 \cite{Behrenz01112001} of 785 vacancies in Sweden, there were on average 20 applicants for each opening. In this study, the employers provided their screening steps and reasons for applicant elimination. The most common reasons were insufficient education and experience, with 55\% and 63\% respectively citing these factors in the questionnaire.  
Notably, 21\% of employers cited age (over 45) as the reason, while 6.5\% indicated it was due to the applicant being a woman aged between 20-30.

Recent audit studies also find that older applicants are less likely to receive callbacks \cite{Lahey2008,Farber2019,OlderWorkersFieldExperiment}. These studies often also find that older women face more age discrimination than older men \cite{Burn2020b}, such as twice as much in retail sales \cite{OlderWorkersFieldExperiment}.

There is extensive research on the various types of bias that exists in the traditional job market \cite{Lippens2023}. This bias is not singular in type, and is often subtle, and some of the bias is accidental, intentional, preconceived, or even hidden \cite{Jones_Arena_Nittrouer_Alonso_Lindsey_2017}. 
In some cases, bias can emerge through unintentional means -- for example, when a high number of resumes meet the application criteria, providing equal time and consideration to each individual submission becomes challenging. To preserve time, a set of criteria unrelated to the merit or skill of applicants might be used, intentionally or unconsciously, to select resumes to advance to the callback stage \cite{ethnicbias}. However, even minor instances of subgroup bias can lead to significant levels of hiring discrimination \cite{doi:10.1177/0149206320982654}. Due to the labor intensive process of resume screening, some companies have begun automating initial screening steps using artificial intelligence. However, these processes pose challenges, particularly in defining fair and appropriate criteria for selecting and advancing applicants. Various laws prohibit discrimination based on race, color, religion, sex (including gender identity, sexual orientation, and pregnancy), national origin, age (usually 40 or older), disability, or genetic information. Yet, removing discrimination becomes more challenging as systems grow in complexity. Moreover, studies show mixed perceptions of hiring algorithms, with many viewing them as less fair.
\cite{https://doi.org/10.1111/ijsa.12246,automatedinterviews,NEWMAN2020149}.

Generally, companies using machine learning use resume screening for detecting keywords, patterns, and traits \cite{BoRi18a}. Resumes that pass the automated filter are reviewed by a human, who then selects candidates for callbacks. Hence, the key metric for automation is ensuring that the relevant resumes advance to human review.

\subsection{Label bias and algorithmic fairness}

Within the extensive literature on algorithmic fairness~\cite{barocas-hardt-narayanan}, there is an emerging line of work focusing on the issue of \textit{label bias} -- a recognition that in many domains the human decisions, which serve as ground truth labels for training and evaluating models, are themselves influenced by bias that must be accounted for. \citet{fish2016confidence} were among the earliest to note the problem of label bias in algorithmic fairness, which they explore via simulation studies. 

Building on this work, \citet{NEURIPS2019_373e4c5d} challenge the assumption that there is an inherent trade-off to be made between fairness and accuracy. They propose that when biases such as selection and label bias are accounted for, the trade-off between accuracy and fairness can diminish or even disappear. To support this claim, they propose evaluating fairness in a situation where unbiased ground truth labels are available. Additionally, they introduce a semi-supervised learning approach that leverages both the unlabeled data and the fairness constraints. As in \citet{fish2016confidence}, simulation studies are required to explore data that does not have selection or label bias.

Similarly, \citet{DBLP:journals/corr/abs-2102-03054} focus on identifying and removing training instances affected by label bias in historical datasets. To identify such instances, they find matched pairs of instances that receive different labels, and assume that the one with the least confident classification decision is the one that has received discrimination. As above, this work relies on synthetically generated instances for training and testing.

Finally, \citet{jiang2020identifying} present a mathematical framework for mitigating label bias by assuming that there is an existing underlying unbiased label function. They introduce a re-weighting scheme that adjusts the significance of some training instances to account for label bias. However, this approach is generally designed for fairness metrics that do not require unbiased labels (e.g., demographic parity~\cite{dwork2012fairness}). For other measures,  additional assumptions and estimates are required.

In the context of this related work, we offer the following contributions:
\begin{itemize}
    \item By using audit study data to analyze algorithmic fairness, we are able to control for selection bias as well as rigorously quantify the amount of label bias present.
    \item We introduce a new method to correct for label bias using individual treatment effect (ITE) estimators.
    \item We provide empirical evidence demonstrating that traditional fairness evaluation metrics, when applied to conventionally biased labels, may produce misleading conclusions about algorithmic fairness.
\end{itemize}

\section{Training and Evaluating Classifiers using Human Audit Study Data}
\label{sec:data}
In this section, we introduce the audit study dataset used in our evaluations, then describe how we adapt this dataset to train and evaluate a resume screening machine learning model based on standard classifiers from the machine learning literature \cite{pedregosa2011scikit}.

\subsection{Human Audit Study Data}
Our data is drawn from a large scale field experiment investigating age discrimination in hiring~\cite{OlderWorkersFieldExperiment}. Over 40,000 synthetic resumes were created and sent in response to online job postings for four types of entry level positions: retail sales, administrative assistants, janitors, and security. The primary outcome of interest was whether the synthetic applicant received a callback from the potential employer. The aim of the study was to determine whether callback rate was influenced by age, all else being equal, comparing young (age 29-31), middle age (age 49-51), and old (age 64-66) applicants.

\begin{table}[t!]
\centering
    \begin{tabular}{ | p{2cm} | p{5.4cm} | }
    \hline
    \textbf{Variable} & \textbf{Description} \\ \hline
    city-zip & The zipcode and city of the applicant \\ \hline
    age group & Applicant’s age group: young (29–31), middle (49–51), or old (64–66). \\ \hline
    gender & Applicant gender: Female or Male. \\ \hline
    employment & Whether the applicant is currently employed. \\ \hline
    occupation & Occupation of the job, one of \{Admin, Sales, Janitor, Security\}. \\ \hline
    type & Resume type: Y (Young), M (Middle), O (Old), B (Bridging), BL (Late bridging), BE (Early bridging). \\ \hline
    template & Word template used for the resume, one of \{A, B, C\}. \\ \hline
    spanish & Whether Spanish is listed as a skill. \\ \hline
    internship & Whether the position applied for is an internship. \\ \hline
    customer ~~~~ service & Whether customer service experience is listed. \\ \hline
    cpr & Whether CPR training is listed. \\ \hline
    tech skills & Whether technical skills (e.g., Office, POS systems) are listed. \\ \hline
    wpm & Application mentions typing speed in words per minute (45, 50, or 55; integer). \\ \hline
    grammar & Whether the application did not include two typos. \\ \hline
    college & Whether the applicant attended college. \\ \hline
    employee month & Whether the applicant won an employee of the month award. \\ \hline
    volunteer & Whether volunteer experience is listed. \\ \hline
    skill & Skill level: high (1) or low (0). \\ \hline
    \end{tabular}
\caption{List of features used in the callback classifier. }
\label{table:data_description}
\end{table}

As the goal of such human audit studies is to rigorously estimate potential discrimination, considerable care is given to the manner in which resumes are created. In particular, resumes of comparable skill and experience were sent to the same ad -- varying only the age of the applicant -- in order to isolate the effect of age on callback.\footnote{There are many nuances here (e.g., older applicants should have longer work histories, on average); please see \citet{OlderWorkersFieldExperiment} for more details.}

The field study found strong overall evidence of age discrimination, with callback rates significantly lower for middle-aged ($\downarrow 18\%$) and older ($\downarrow 35\%$ ) applicants, as compared to younger applicants, despite having comparable resumes.

\subsection{Training Classifiers on Human Audit Study Data}

Given the data described above, we aim to study the behavior of a machine learning system trained to replicate the human decisions in the data. This simulates a scenario in which a firm attempts to automate the callback decision-making process based on historical decisions of its staff. As \citet{OlderWorkersFieldExperiment} rigorously show, there is considerable discrimination present in this data; we wish to study its effect on the fairness and accuracy of resulting classifiers, as well as our ability to measure these values.


To do so, we train various classifiers to predict the binary callback variable $Y \in \{0,1\}$ given applicant attributes $\X \in \mathbb{R}^d$, as well as the protected age attribute $A \in \{y,o\}$, where $y$ indicates younger applicants and $o$ indicates older applicants.\footnote{For simplicity, we collapse middle and older applicants into a single age group called ``older'', with the remainder comprising the ``younger'' group.} Applicant attributes include demographics (gender, location), employment status, foreign language skills, typing speed, college experience, and volunteering history, see Table~\ref{table:data_description} for more details on the variables used in our study.

\begin{table}[t]
    \centering
    \begin{tabular}{|l|r|r|r|}
        \hline
        \textbf{Age Group} & \textbf{Callback} & \textbf{No Callback} & \textbf{Total} \\
        \hline
        Young & 2,505 (19\%) & 10,896 (81\%) & 13,401 \\
        Old/Middle & 3,587 (14\%) & 21,945 (86\%) & 25,532 \\
        \hline
        Total & 6,095 (16\%) & 32,892 (84\%) & 38,987 \\
        \hline
    \end{tabular}
    \caption{Callback data by age group.}
    \label{tab:callback_agegroup}
\end{table}

We use the human audit study data as a labeled dataset $D = \{(\X_1, A_1, Y_1), \ldots, (\X_n, A_n, Y_n)\}$. Table~\ref{tab:callback_agegroup} displays the callback rates by age group, showing a roughly 5\% greater callback rate for younger applicants over older applicants. We experiment with two standard classification models, random forests and neural networks, performing cross-validation to evaluate fairness and accuracy of the models.

\subsection{Evaluation Measures}

To measure accuracy while accounting for class imbalance, we use Area Under the ROC Curve (\textbf{AUC})~\cite{fan2006understanding}. To measure fairness, for simplicity and clarity we focus primarily on False Positive Rate Disparity (\textbf{FPRD}). FPRD is the difference in false positive classification rates between young and old applicants:
$$FPRD = FPR_y - FPR_o$$
False positive rates for each age group are defined as usual:
$$FPR_y = \frac{FP_y}{FP_y + TN_y}, \hspace{1em} FPR_o = \frac{FP_o}{FP_o + TN_o}$$ 
where $FP$ (\textit{false positives}) is the number of negative instances incorrectly classified as positive, $TN$ (\textit{true negatives}) is the number of negative instances correctly classified as negative, and $FP+TN$ is the total number of \textit{actual negative} instances ($AN$). 

FPRD will be \emph{positive} when the classifier discriminates against older applicants, \emph{negative} when it discriminates against younger applicants, and \emph{zero} when the classifier does not discriminate based on age.  Within the hiring setting, a false positive can be viewed as an applicant receiving a callback when they should not have. Thus, if FPRD is positive, then the rate at which unqualified younger applicants receive callbacks is higher than that of unqualified older applicants. 

To ensure reliable comparison across classifiers with different positive prediction rates, we standardize the number of predicted positives by fixing a common callback budget. Specifically, since the original dataset has an overall callback rate of 16\%, we enforce the same rate during evaluation. For each classifier, we sort the test instances by their predicted probability of receiving a callback and label the top 16\% as predicted positive.

\section{Label Bias and the Illusion of Fairness}
\label{sec:illusion}
The primary problem in using human-provided labels derived from the field experiment to train and evaluate a classifier is the presence of \textit{label bias}~\cite{NEURIPS2019_373e4c5d,jiang2020identifying,DBLP:journals/corr/abs-2102-03054}. That is, for each applicant $i$ we only observe the label $y_i$, which we know is the result of a decision-making process measurably influenced by age discrimination. Unfortunately, we cannot observe the idealized label $y_i^*$ that would result from a process free of discrimination.

This label bias will corrupt our measures of both accuracy and fairness.  A classifier that prefers younger applicants to older applicants may appear more accurate, as it will better reflect human-generated labels. Likewise, measures of discrimination such as FPRD may be underestimated when computed using data with label bias. This is because removing label bias results in fewer young applicants getting callbacks and more older applicants getting callbacks, altering the false positive rates of each group.

Table~\ref{tab:example} shows a simple example of how removing label bias can alter FPRD. Initially, both groups have the same number of actual negatives (AN), though the classifier has a 10 percentage point higher false positive rate for younger applicants. To remove label bias, we assume that 20 younger applicants who were given callbacks in the human audit study should not have been; of these, 15 were labeled as callbacks by the classifier. As a result, for the younger group $AN_y$ increases to 120, while $FP_y$ increases to 45, resulting in an increase in $FPR_y$ from $0.3 \rightarrow 0.375$. Conversely, for the older group  $AN_o$ decreases to 80, while $FP_o$ decreases to 5, resulting in a decrease in $FPR_o$ from $0.2 \rightarrow 0.0625$. Thus, in this example, removing label bias affecting 40 applicants increased the FPRD estimate from $0.1 \rightarrow 0.3125$. 

\begin{table}[t]
\centering
\begin{tabular}{|c|c|c|c|c|}
\hline
\textbf{Group} & \textbf{AN} & \textbf{FP} & \textbf{FPR} & \textbf{FPRD} \\
\hline
Y (Before) & 100 & 30 & 0.3 & \multirow{2}{*}{\textbf{0.1}}\\
O (Before)   & 100 & 20 & 0.2 & \\
\hline
Y (↓  callbacks) & 120 & 45 & 0.375 & \multirow{2}{*}{\textbf{0.3125}}\\
O (↑  callbacks)   & 80  & 5 & 0.0625  &  \\
\hline
\end{tabular}
\caption{Illustrative example of how reducing label bias --
increasing true callbacks (reducing actual negatives) for older applicants and decreasing true callbacks (increasing actual negatives) for younger applicants -- can affect False Positive Rate Disparity (FPRD).}
\label{tab:example}
\end{table}

This example illustrates how sensitive fairness metrics are to the presence of label bias, and underscores the need to more rigorously assess the presence of discrimination in labeled data. Otherwise, label bias can create an illusion of fairness, causing classifiers to appear less discriminatory than they truly are.

\section{Repairing Label Bias with Individual Treatment Effect Estimation}
\label{sec:ite}
The preceding discussion suggests a path forward. If we were able to remove label bias, we would not only improve the reliability of our fairness measures, but also create cleaner training data to fit the classifier in the first place. However, doing so raises two questions: (1) which applicants should have their callback labels amended? and (2) how many labels do we need to amend?

Typical samples of convenience make these questions difficult to answer. For example, if in the real-world application pool younger applicants tend to be more qualified than older applicants, then it is not clear what the true callback rates should be for each group. However, this is where the value of human audit studies arises. By design, the human audit study expects equal callback rates between the two age groups. Indeed, this was the motivation for carefully controlling for other resume attributes when constructing the synthetic resumes. Thus, the answer to our second question is: amend the labeled data until the callback rates are equal for younger and older applicants.

As for the first question -- which labels to amend -- our approach is based on the intuition that applicants for which age was a determinative factor in the callback decision are those that should have their callback decisions amended. In other words, for younger applicants who received a callback, if they had been older applicants, would they still have received a callback? Analogously, for older applicants who did not receive a callback, if they had been younger, would they have received a callback after all? 

Framed in this way, this becomes a counterfactual question -- what would the callback outcome have been for an applicant if they had been in a different age group? To answer this question, we build on a long line of work in the social and medical sciences on \textit{individual treatment effect estimation}.


Given the human audit study dataset $D = \{(\X_1, A_1, Y_1), \ldots, (\X_n, A_n, Y_n)\}$, we treat the age variable $A_i$ as a binary {\it treatment} indicator representing whether $i$ is in the younger treatment ($A_i=1$) group or older control ($A_i=0$) group, and $Y_i$ is the observed callback {\it outcome} for individual $i$.  We are interested in quantifying the causal effect that treatment $A$ has on the outcome $Y$. The fundamental problem of causal inference is that we can only observe one outcome per individual, either the outcome of an individual receiving a treatment or not. Thus, we do not have direct evidence of what might have happened had we given individual $i$ a different treatment. 

Rubin's potential outcome framework is a common way to formalize this fundamental problem~\cite{rubin1974estimating}. Let $Y^{(1)}$ indicate the potential outcome an individual would have gotten had they received treatment $(A=1)$, and similarly let $Y^{(0)}$ indicate the outcome an individual  would have gotten had they received no treatment ($A=0$). While we cannot observe both $Y^{(1)}$ and $Y^{(0)}$ for the same individual, we can now formally express the quantity of interest. We are interested in the {\sl Individual Treatment Effect} (ITE), which is the expected difference in outcome for a specific type of individual:
\begin{align}
\tau(\x) = \mathop{\mathbb{E}}[Y^{(1)} | \X=\x] -   \mathop{\mathbb{E}}[Y^{(0)} | \X=\x]
\label{eq:ite}
\end{align}
that is, the treatment effect for individuals where $\mathbf{X}=\mathbf{x}$. For example, if the covariate vector represents the (gender, height) of a person, then the ITE will estimate treatment effects for individuals that match along those variables.

Using standard assumptions, we can estimate ITE as follows:
\begin{align}
\hat{\tau}(\mathbf{x}) = & \mathop{\mathbb{E}}[Y|A=1, \mathbf{X}=\mathbf{x}] - \mathop{\mathbb{E}}[Y|A=0, \mathbf{X}=\mathbf{x}]\\
\label{eq:ite_estimator}
= & \frac{1}{|S_1(\mathbf{x})|} \sum_{i \in S_1(\mathbf{x})} Y_i -\frac{1}{|S_0(\mathbf{x})|}\sum_{i \in S_0(\mathbf{x})} Y_i
\end{align}
where $S_1(\mathbf{x})$ is the set of individuals $i$ such that $\mathbf{X}_i=\mathbf{x}$ and $A_i=1$, and similarly for $S_0(\mathbf{x})$. In other words, Equation~\eqref{eq:ite_estimator} simply computes, for all individuals with covariates equal to $\mathbf{x}$, the difference between the average outcome for individuals in the treatment group and the average outcome for individuals in the control group. For example, if $\mathbf{X=x}$ indicates individuals with (gender=male, height=5), $A=1$ indicates that an individual is in the younger group, and $A=0$ that they are in the older group, then $\hat{\tau}(\hbox{x})$ is the difference in average outcome between individuals who are in the young group and those who are not.

A key challenge to using Equation~\eqref{eq:ite_estimator} in practice is that $\X$ may be high dimensional, leading to a small sample where $\mathbf{X=x}$. In the extreme case, there may be exactly one instance where $\mathbf{X=x}$.  Instead, we adopt the virtual twins approach \cite{Foster@2011subgroup}, a two step procedure to estimate ITE. First, it fits a random forest with all  data (including control samples and treatment samples), where each data is represented by inputs $(\X_i, A_i)$ and outcome $Y_i$. Then, to estimate the ITE for instance $i$, it computes the difference between the predicted values for treatment input $(\X_i, A_i=1)$ and control input $(\X_i, A_i=0)$. The name ``virtual twin'' derives from the fact that for each control input $(\X_i, A_i=0)$, we make a copy $(\X_i, A_i=1)$ as treatment input that is alike in every way to the control input except for the treatment variable. Similarly, for each treatment input $(\X_i, A_i=1)$, we make a copy $(\X_i, A_i=0)$. 

If $\hat{Y}(\x_i,1)$ is the posterior probability of callback ($Y_i=1$) produced by the random forest for input $(\X_i=\x_i, A_i=1)$, and $\hat{Y}(\x_i,0)$ is similarly the probability of callback for input $(\X_i=\x_i, A_i=0)$, then the virtual twin ITE estimate for instance $i$ is:
\begin{align}
\label{def.vt}
\hat{\tau}(\x_i)=\hat{Y}(\x_i,1) - \hat{Y}(\x_i,0)
\end{align}
Thus, $\hat{\tau}(\x_i)$ is the increase in probability of callback attributable to being in the younger applicant group. 

Our final label repair algorithm, then, is given in Algorithm~\ref{alg:ite}. After first computing all ITE estimates, we iteratively change the callback labels most likely influenced by discrimination. At each iteration, we assign one younger applicant with a callback to have a no-callback label, and one older applicant with a no-callback label to have a callback label. We repeat until callback rates are equal between age groups, as expected in a world with no  discrimination.

\begin{algorithm}[t]
\caption{Repairing label bias with ITE}
\begin{algorithmic}[1]
\State Fit a random forest classifier on training data $D_{\text{train}}$
\State Compute age ITE estimates $\hat{\tau}(\mathbf{x}_i)$ for each instance in the test set $D_{\text{test}}$
\While{callback rate is not equal between age groups}
    \State \# Flip positive to negative for younger group
    \State Find $i$ with largest $\hat{\tau}(\mathbf{x}_i)$ where $Y_i = 1$, $A_i = 1$
    \State Set $Y_i \gets 0$
    \State \# Flip negative to positive for older group
    \State Find $j$ with smallest $\hat{\tau}(\mathbf{x}_j)$ where $Y_j = 0$, $A_j = 0$
    \State Set $Y_j \gets 1$
\EndWhile
\end{algorithmic}
\label{alg:ite}
\end{algorithm}

\section{Experiments}
\label{sec:experiments}
The goal of our experiments is to answer the following questions:

\begin{itemize}
    \item \rqone
    \item \rqtwo 
     \item \rqthree
    \item \rqfour
\end{itemize}
We consider three types of data pre-processing:
\begin{enumerate}
    \item \textbf{Base Rate (BR)}: Do not perform any manipulation of either the training or testing data.
    \item \textbf{Equalized Base Rate (EBR)}: A simple yet common pre-processing technique to improve fairness is to downsample data to ensure equal class distributions across protected classes~\cite{kleinberg2016inherent,li2022data}. To do so, we  delete from the training set at random older applicants who did not get a callback until the callback rates are equal for the two age groups.
    \item \textbf{Individual Treatment Effect Repair (ITE)}: The method from Algorithm~\ref{alg:ite}.
\end{enumerate}

To further investigate  the impacts of these interventions separately on the training and testing data, we consider the following settings:
\begin{itemize}
    \item \textbf{ITE Train \& Test:} Apply ITE to both training and testing data. To do so, we perform cross-validation, estimating ITE for the test set in each fold. Then, for each fold, we apply Algorithm~\ref{alg:ite} separately for the train and test sets, ensuring equal callback rates between groups in both sets.
    \item \textbf{EBR Train - ITE Test:} Apply EBR to the training data, but ITE to the testing data. 
\end{itemize}
Comparing EBR with (EBR Train - ITE Test) allows us to isolate the effect of label bias in the testing data on fairness measures.

We consider two classification models: Random Forest and Multi-Layer Perceptron. For the Random Forest, key parameters are: number of estimators=50, minimum samples per split=2, minimum samples per leaf=1. For Multi-Layer Perceptron, we use three hidden layers of sizes (128, 64, 32), ReLU activation functions, and the Adam optimizer. All experiments are conducted in scikit-learn~\cite{pedregosa2011scikit}.

\subsection{Simulating Selection Bias}
\label{sec:sim}
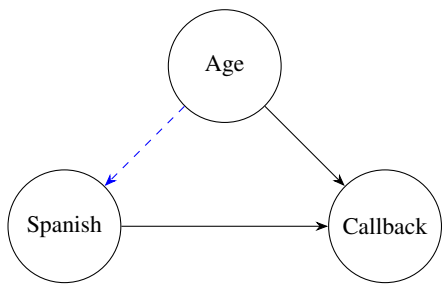
\begin{figure}[t]
  \centering
  \small
\begin{tikzpicture}[
    >=Stealth,
    node distance=1.5cm, 
    every node/.style={draw, circle, minimum size=1.5cm}
  ]
  \node (Age)       {Age};
  \node (Spanish) [below left=of Age]    {Spanish};
  \node (Callback) [below right=of Age]  {Callback};

  \draw[->, dashed, blue] (Age)     -- (Spanish);
  \draw[->] (Age)     -- (Callback);
  \draw[->] (Spanish) -- (Callback);
\end{tikzpicture}
\caption{Causal diagram depicting our approach to injecting selection bias in audit study data (\S\ref{sec:sim}).}
  \label{fig:spanish}
\end{figure}

\begin{figure*}[t]
  \centering
  \begin{subfigure}[t]{0.48\textwidth}
    \centering
    \includegraphics[width=\linewidth]{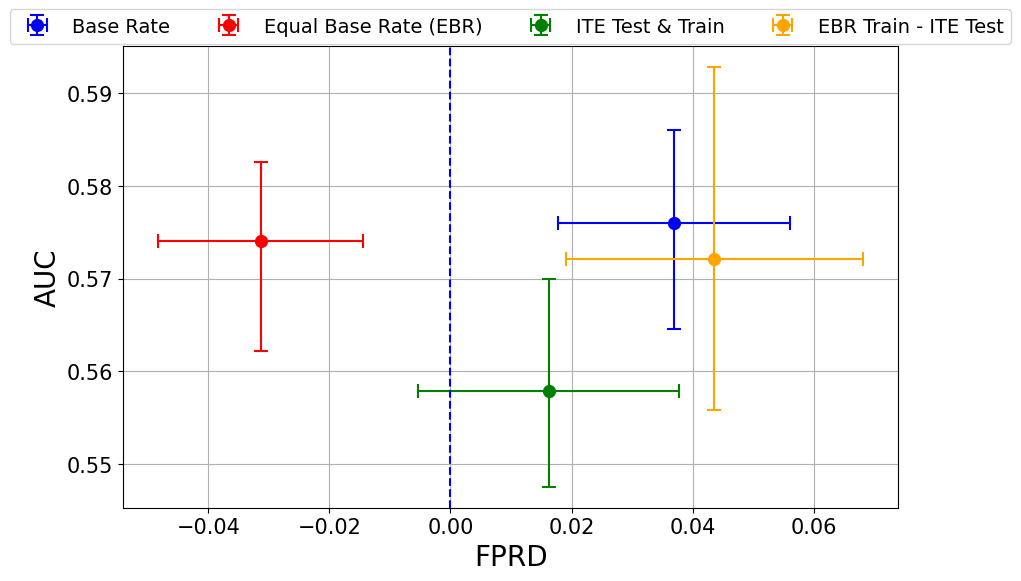}
    \caption{Random Forest }
    \label{fig:auc_vs_fprd_rf}
  \end{subfigure}\hfill
  \begin{subfigure}[t]{0.48\textwidth}
    \centering
    \includegraphics[width=\linewidth]{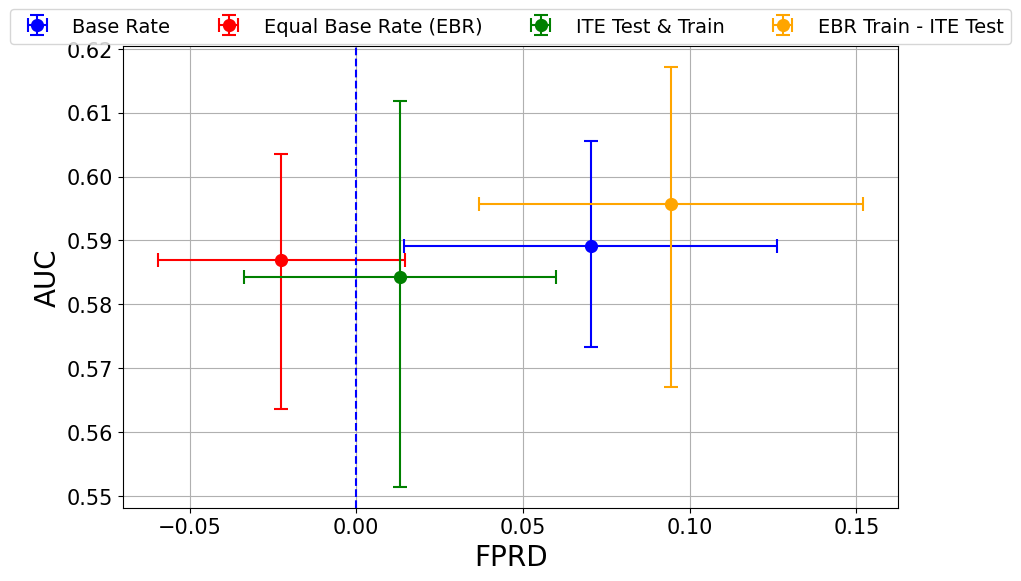}
    \caption{Neural Network}
    \label{fig:auc_vs_fprd_nn}
  \end{subfigure}
  \caption{Comparison of accuracy (AUC) and fairness (FPRD) across different approaches to handle label bias. Positive FPRD indicates discrimination against older applicants.}
  \label{fig:auc_vs_fprd}
\end{figure*}

\begin{figure*}[t]
  \centering
  \begin{subfigure}[t]{0.48\textwidth}
    \centering
    \includegraphics[width=\linewidth]{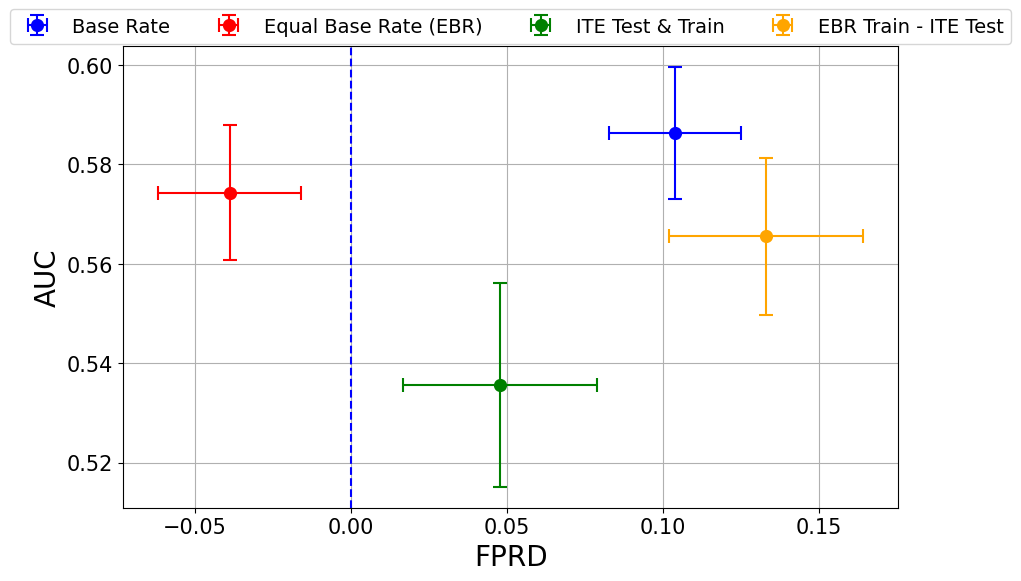}
    \caption{Random Forest}
    \label{fig:auc_vs_fprd_rf_double}
  \end{subfigure}\hfill
  \begin{subfigure}[t]{0.48\textwidth}
    \centering
    \includegraphics[width=\linewidth]{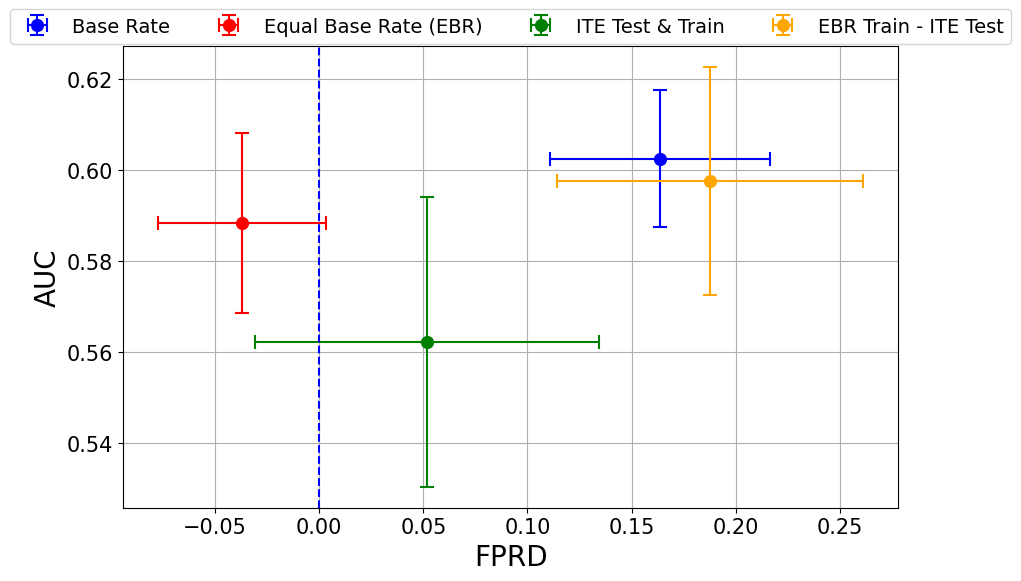}
    \caption{Neural Network}
    \label{fig:auc_vs_fprd_nn_double}
  \end{subfigure}
  \caption{Comparison of accuracy (AUC) and fairness (FPRD) across different approaches to handle label bias. The original human discrimination in the audit data is doubled from that of Figure~\ref{fig:auc_vs_fprd}. Positive FPRD indicates discrimination against older applicants. }
  \label{fig:auc_vs_fprd_double}
\end{figure*}

In order to investigate RQ4, we need to generate data that reflects the typical non-audit study data. That is, we must reintroduce the sample selection bias that pervades samples of convenience normally used to train classifiers to automate human decision making. Selection bias occurs when the distribution of data inadvertently introduces undesired correlations between the features pertaining to a protected attribute and the class label. For example, suppose that Spanish fluency is predictive of callbacks, and that younger applicants are more likely to speak Spanish. Figure~\ref{fig:spanish} displays the causal model for such a scenario. In this case, the selection bias of Spanish may introduce an unintended relationship between age and callback. By contrast, in the audit study data, resumes were generated specifically such that Spanish fluency was not correlated with age, thereby avoiding this issue.

To study the impact of this selection bias, we conduct additional experiments where we sample the original audit study data in order to vary the correlation between age and Spanish fluency, while holding constant the relationship between Spanish fluency and callbacks. For example, to increase the prevalence of Spanish among younger applicants, we drop at random older applicants with Spanish and younger applicants without Spanish, while also keeping the callback rate of Spanish applicants constant. We consider a range of values for the conditional probabilities $P(\hbox{Spanish}=1|A=y)$ and $P(\hbox{Spanish}=1|A=o)$ to investigate how the disparity in Spanish fluency by age influences algorithm behavior.


\section{Results}
\label{sec:results}
In this section, we present results investigating each of the four research questions.

\begin{figure*}[t]
  \centering
  \begin{subfigure}[t]{0.48\textwidth}
    \centering
    \includegraphics[width=\linewidth]{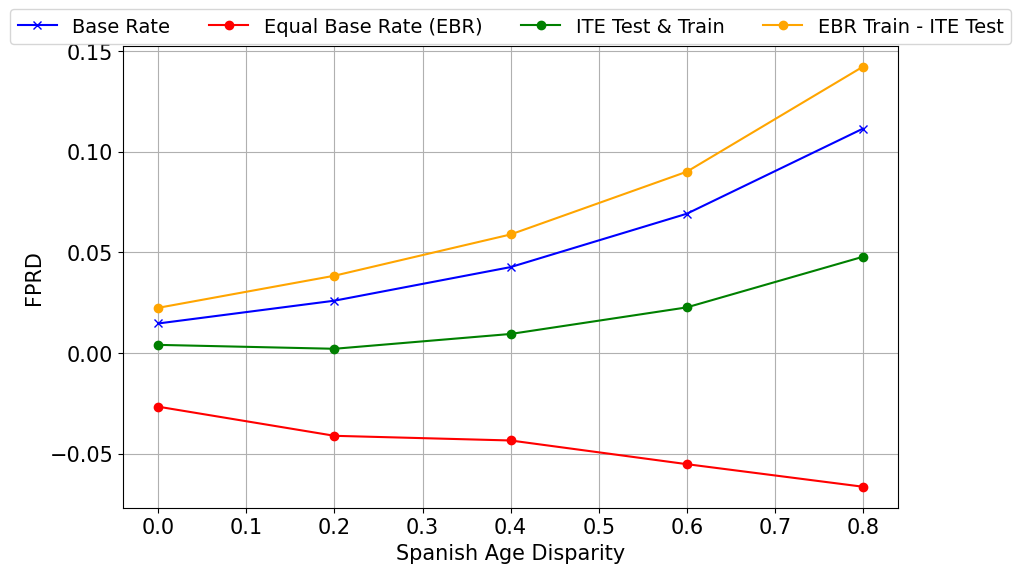}
    \caption{Random Forest}
    \label{fig:fprd_vs_spanish_rf}
  \end{subfigure}\hfill
  \begin{subfigure}[t]{0.48\textwidth}
    \centering
    \includegraphics[width=\linewidth]{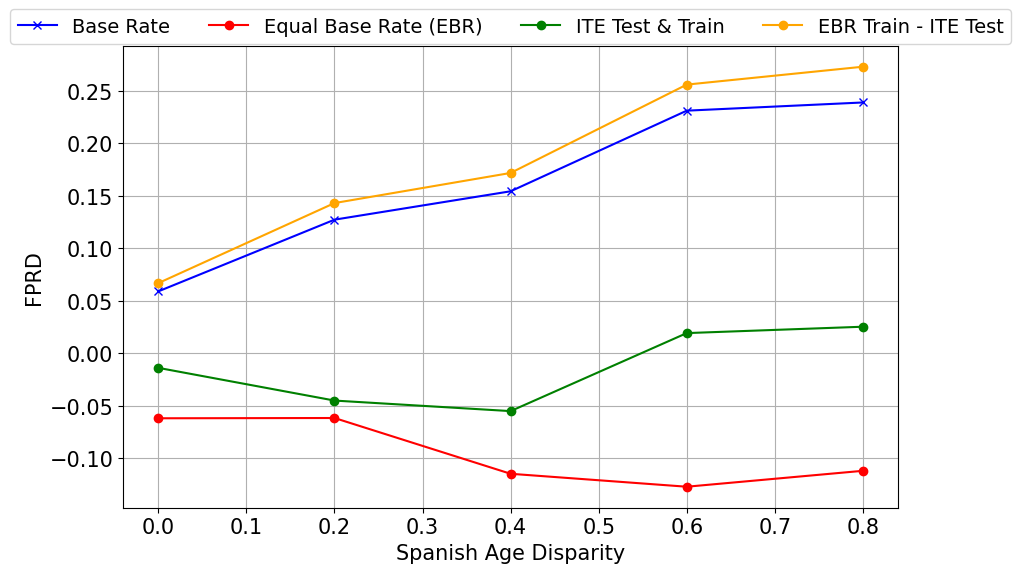}
    \caption{Neural Network}
    \label{fig:fprd_vs_spanish_nn}
  \end{subfigure}
  \caption{Fairness comparison as selection bias due to Spanish fluency varies by age group. The $x$-axis, Spanish Age Disparity, is equal to $P(\hbox{Spanish}=1|A=y)-P(\hbox{Spanish}=1|A=o)$. Thus, larger $x$ values indicate that Spanish fluency is more prevalent for younger applicants.}
  \label{fig:fprd_vs_spanish}
\end{figure*}

\textbf{RQ1: Effect of label bias in test data. }
Figure~\ref{fig:auc_vs_fprd} displays the main AUC versus FPRD results for our primary methods, with mean and standard deviation computed from 5-fold cross-validation. We first focus on the effects of debiasing the test data with ITE. For random forest (Figure~\ref{fig:auc_vs_fprd_rf}), the Base Rate has FPRD $\approx 0.038$, exhibiting similar discrimination against older applicants as observed in the original audit study dataset. Applying the Equal Base Rate intervention, and evaluating on the unmodified test set, at first appears to have removed the discrimination against over applicants. Indeed, EBR seems to have over-adjusted, now showing a slight preference for older applicants (FPRD $\approx -0.031$). However, we observe a dramatic difference when we repair the label bias in the test set using ITE. EBR Train - ITE Test shows that the discrimination against older applicants remains, even after equalizing base rates, and indeed the sign of FPRD changes when label bias is removed (FPRD $\approx 0.042$ vs. $-.031$ for EBR). This discrepancy of 0.073 in FPRD suggests that label bias can have dramatic impacts on estimates of fairness, which has broader implications on our ability to measure fairness.

For neural networks  (Figure~\ref{fig:auc_vs_fprd_rf}), the relative comparisons between the methods are largely similar. Noticeable differences are a somewhat higher AUC overall compared to random forest (e.g., Base Rate $0.576 \rightarrow 0.59$), as well as much large values for FPRD. For example, EBR Train - ITE Test increases FPRD from $0.042$ using random forests to $0.095$ using neural networks. This suggests that models with more degrees of freedom may be even more susceptible to label bias. Thus, in answering RQ1, we see that simply employing EBR can give the illusion of fairness but in reality still exhibits nearly 10\% age disparity in its decisions.

\textbf{RQ2: Effect of label bias in training data.} Continuing the discussion of Figure~\ref{fig:auc_vs_fprd}, we next compare the result of ITE Test \& Train, which performs the ITE label debiasing to both the training and test set. We observe that this results in the lowest amount of discrimination (FPRD $\approx 0.017$), though this does coincide with a roughly 1\% decrease in AUC. 

When compared with EBR Train - ITE Test, we see that ITE Test \& Train reduces FPRD from 0.042 $\rightarrow$ 0.017, a 60\% reduction in disparity. Even though EBR ensures that the two age groups receive equal callback rates in the training data, the label bias remains. Thus, under-qualified younger applicants in the training data receive callbacks at higher rates than under-qualified older applicants. And, conversely, qualified older applicants are more likely to not receive a callback than qualified younger applicants. By learning these patterns, the resulting classifier replicates this discrimination in the test set.

\textbf{RQ3: Effect of label bias magnitude.} 
To investigate the impact of the amount of human discrimination in the original audit data, we create a modified version of the data with more discrimination than the original. Specifically, we sample the original data by removing at random older applicants who received a callback until the callback difference between age groups increases to 10\%, roughly double the original difference. Figure~\ref{fig:auc_vs_fprd_double} shows the results for all methods. We can see that the overall patterns remain. Critically, we observe that the discrepancies in evaluation measures exhibit a commensurate increase in severity due to the doubling of discrimination. For example, for random forest (Figure~\ref{fig:auc_vs_fprd_rf_double}), EBR has FPRD $\approx -.038$, compared to an FPRD $\approx 0.133$ for EBR Train-ITE Test. This is a discrepancy of 0.171, more than double the discrepancy of 0.073 in the original results in Figure~\ref{fig:auc_vs_fprd_rf}. Once again, without adjusting for label bias, the sign of the discrimination is estimated incorrectly. These results suggest that label bias can be more detrimental precisely in domains with large amounts of discrimination.

Encouragingly, ITE Test \& Train appears to maintain only modest discrimination even in this more extreme setting  (FPRD $\approx 0.049$ vs. $0.017$ in the original), which is also observed in the neural network results (Figure~\ref{fig:auc_vs_fprd_nn_double}). While discrimination does grow, the lower starting point suggests the ITE approach can be more robust to higher levels of discrimination in the audit data. Unfortunately, this does appear to come at a decrease in AUC of roughly 2.1\%. Further research is needed to understand this interplay between discrimination in audit data, fairness, and accuracy.

\textbf{RQ4: Effect of selection bias.}
Finally, we analyze how results vary when we reintroduce selection bias into the dataset. Figure~\ref{fig:fprd_vs_spanish} plots FPRD when we vary the disparity in Spanish fluency by age group.  The x-axis, $P(\hbox{Spanish}=1|A=y)-P(\hbox{Spanish}=1|A=o)$, represents how much more likely Spanish fluency is among younger applicants than older applicants. For example, when $x=0.8$, $P(\hbox{Spanish}=1|A=y)=0.9$ and $P(\hbox{Spanish}=1|A=o)=.1$. That is, 90\% of younger applicants in that sample speak Spanish, while only 10\% of older applicants do. While we keep callback rates constant across age groups otherwise, because Spanish fluency is a desirable property, we see greater discrimination (larger FPRD) against older applicants as data contains relatively fewer older applicants who speak Spanish.

Similar to Figure~\ref{fig:auc_vs_fprd}, we observe substantial and significant differences in EBR depending on whether the test set has been debiased. This is most pronounced when Spanish Age Disparity is $0.8$, where for EBR FPRD $\approx -.08$ and for EBR - ITE Test FPRD $\approx 0.145$ for random forest (Figure~\ref{fig:fprd_vs_spanish_rf}). The discrepancy is even larger for the neural network, which shows a FPRD discrepancy of 0.38 ($-0.11 \rightarrow 0.27$, Figure~\ref{fig:fprd_vs_spanish_nn}). Here again, not only is the magnitude of the difference extreme, but the direction of discrimination changes from against younger applicants to against older applicants. E.g., for the neural network, two models trained in identical fashion can be found to either have a 0.11 discrimination towards older applicants, or a 0.27 discrimination towards younger applicants, depending on whether or not the test data has been repaired of label bias. 

Additionally, we find that although the ITE Test \& Train approach consistently achieves the lowest disparity among methods evaluated on debiased test sets, it still begins to exhibit a bias in favor of younger applicants under the most extreme conditions (e.g., at a bias level of 0.8). This further reinforces the value of human audit study data and highlights the limitations of relying solely on statistical corrections. By randomizing both covariates and protected attributes, audit studies offer a cleaner and more controlled source of training data that can mitigate the influence of sampling bias in algorithmic decision making.

\section{Discussion}
\label{sec:discussion}
These results have broader implications for the design, evaluation, and deployment of fairness interventions in automated decision support systems. Our findings demonstrate that achieving parity in observed outcomes -- such as equal callback rates across groups -- is not, by itself, a sufficient pre-processing intervention to ensure fairness. In particular, interventions that equalize base rates without accounting for  label bias may create the illusion of fairness while leaving meaningful disparities unaddressed.

This has consequences for both researchers and practitioners. Simply selecting or resampling data to ensure equal callback rates can mask underlying discrimination, especially when the original labels reflect biased human decisions. A system that appears fair in terms of aggregate outcomes may still produce systematically different errors across groups, such as higher false positive rates for one demographic. This underscores the need for fairness metrics and interventions that go beyond surface-level parity and consider the causal mechanisms driving observed disparities.

More broadly, our results highlight the importance of using richer sources of evaluation data -- such as audit studies or experimental designs -- that enable more precise identification of when and how discrimination occurs. Without access to such data, fairness assessments risk being overly reliant on biased labels and incorrect assumptions about group differences.

\subsection{Limitations}

While our study provides a novel approach to evaluating fairness interventions using audit study data and individual treatment effect estimation, several limitations remain. First, despite leveraging audit studies to better approximate real-world discrimination, like all such studies we still lack access to the true underlying ground truth of applicant quality. This limits our ability to definitively assess whether corrected labels fully reflect fair outcomes. Future work should investigate rigorous simulation studies to better understand how robust these approaches are to different distributions of label bias. 

Second, our analysis focuses on a limited set of fairness interventions and measures. While we demonstrate the shortcomings of base rate equalization and introduce an ITE-based alternative, a more comprehensive comparison with other debiasing approaches -- such as adversarial learning, reweighting schemes, or post hoc calibration \cite{pessach2022review} -- would provide a broader understanding of how label bias affects these other techniques.

Finally, we evaluate our methods on a single audit study dataset focused on age discrimination in hiring. Although this domain is high-impact, future work should extend the analysis to additional datasets and settings, including other protected attributes (e.g., race or gender), to assess the generalizability of our findings.

\section{Conclusion and Future Work}
\label{sec:conclusion}

We introduced a novel approach using human audit study data to better measure and mitigate algorithmic fairness in the presence of label bias. Our method leverages  Individual Treatment Effect estimates to assess whether individuals receive fair predictions and repairs the data accordingly. Our empirical results indicate that this approach leads both to fairer predictions can reduce the ``\emph{illusion of fairness}'' of traditional approaches that do not account for  label bias. These results point toward the need for future studies that can efficiently incorporate audit studies into AI-augmented decision making processes.

\bibliography{audit_studies}

\begin{thebibliography}{38}
\providecommand{\natexlab}[1]{#1}

\bibitem[{Bandy(2021)}]{bandy2021algoaudit}
Bandy, J. 2021.
\newblock Problematic Machine Behavior: A Systematic Literature Review of Algorithm Audits.
\newblock \emph{Proc. ACM Hum.-Comput. Interact.}, 5(CSCW1).

\bibitem[{Barocas, Hardt, and Narayanan(2023)}]{barocas-hardt-narayanan}
Barocas, S.; Hardt, M.; and Narayanan, A. 2023.
\newblock \emph{Fairness and Machine Learning: Limitations and Opportunities}.
\newblock MIT Press.

\bibitem[{Barron, Bishop, and Dunkelberg(1985)}]{RePEc:tpr:restat:v:67:y:1985:i:1:p:43-52}
Barron, J.; Bishop, J.; and Dunkelberg, W.~C. 1985.
\newblock Employer Search: The Interviewing and Hiring of New Employees.
\newblock \emph{The Review of Economics and Statistics}, 67(1): 43--52.

\bibitem[{Behrenz(2001)}]{Behrenz01112001}
Behrenz, L. 2001.
\newblock Who Gets the Job and Why? an Explorative Study of Employers'recruitment Behavior.
\newblock \emph{Journal of Applied Economics}, 4(2): 255--278.

\bibitem[{Bogen and Rieke(2018)}]{BoRi18a}
Bogen, M.; and Rieke, A. 2018.
\newblock Help Wanted: {A}n Examination of Hiring Algorithms, Equity, and Bias.
\newblock Technical report, Upturn.

\bibitem[{Burn et~al.(2020)Burn, Button, Figinski, and McLaughlin}]{Burn2020b}
Burn, I.; Button, P.; Figinski, T.~F.; and McLaughlin, J.~S. 2020.
\newblock Why {Retirement}, {Social} {Security}, and {Age} {Discrimination} {Policies} {Need} to {Consider} the {Intersectional} {Experiences} of {Older} {Women}.
\newblock \emph{Public Policy \& Aging Report}, 30(3): 101--106.

\bibitem[{Collins et~al.(2021)Collins, Chong, De~Almeida~Neto, Moles, and Schneider}]{Collins2021}
Collins, J.~C.; Chong, W.~W.; De~Almeida~Neto, A.~C.; Moles, R.~J.; and Schneider, C.~R. 2021.
\newblock The simulated patient method: {Design} and application in health services research.
\newblock \emph{Research in Social and Administrative Pharmacy}, 17(12): 2108--2115.
\newblock Publisher: Elsevier BV.

\bibitem[{Derous and Ryan(2018)}]{ethnicbias}
Derous, E.; and Ryan, A. 2018.
\newblock When your resume is (not) turning you down: Modelling ethnic bias in resume screening.
\newblock \emph{Human Resource Management Journal}, 29.

\bibitem[{Dwork et~al.(2012)Dwork, Hardt, Pitassi, Reingold, and Zemel}]{dwork2012fairness}
Dwork, C.; Hardt, M.; Pitassi, T.; Reingold, O.; and Zemel, R. 2012.
\newblock Fairness through awareness.
\newblock In \emph{Proceedings of the 3rd innovations in theoretical computer science conference}, 214--226.

\bibitem[{Evans et~al.(2015)Evans, Roberts, Keeley, Blossom, Amaro, Garcia, Stough, Canter, Robles, and Reed}]{Evans2015}
Evans, S.~C.; Roberts, M.~C.; Keeley, J.~W.; Blossom, J.~B.; Amaro, C.~M.; Garcia, A.~M.; Stough, C.~O.; Canter, K.~S.; Robles, R.; and Reed, G.~M. 2015.
\newblock Vignette methodologies for studying clinicians’ decision-making: {Validity}, utility, and application in {ICD}-11 field studies.
\newblock \emph{International Journal of Clinical and Health Psychology}, 15(2): 160--170.
\newblock Publisher: Elsevier BV.

\bibitem[{Fan, Upadhye, and Worster(2006)}]{fan2006understanding}
Fan, J.; Upadhye, S.; and Worster, A. 2006.
\newblock Understanding receiver operating characteristic (ROC) curves.
\newblock \emph{Canadian Journal of Emergency Medicine}, 8(1): 19--20.

\bibitem[{Farber, Herbst, and Silverman(2019)}]{Farber2019}
Farber, H.~S.; Herbst, C.~M.; and Silverman, D. 2019.
\newblock Whom {Do} {Employers} {Want}? {The} {Role} of {Recent} {Employment} and {Unemployment} {Status} and {Age}.
\newblock \emph{Journal of Labor Economics}, 37(2): 323--349.

\bibitem[{Fish, Kun, and Lelkes(2016)}]{fish2016confidence}
Fish, B.; Kun, J.; and Lelkes, {\'A}.~D. 2016.
\newblock A confidence-based approach for balancing fairness and accuracy.
\newblock In \emph{Proceedings of the 2016 SIAM international conference on data mining}, 144--152. SIAM.

\bibitem[{Foster, Taylor, and Ruberg(2011)}]{Foster@2011subgroup}
Foster, J.~C.; Taylor, J.~M.; and Ruberg, S.~J. 2011.
\newblock Subgroup identification from randomized clinical trial data.
\newblock \emph{Statistics in Medicine}, 30(24): 2867--2880.

\bibitem[{Gaddis(2018)}]{gaddis2018introduction}
Gaddis, S.~M. 2018.
\newblock An introduction to audit studies in the social sciences.
\newblock In \emph{Audit studies: Behind the scenes with theory, method, and nuance}, 3--44. Springer.

\bibitem[{Hutchinson and Mitchell(2019)}]{hutchinson201950}
Hutchinson, B.; and Mitchell, M. 2019.
\newblock 50 years of test (un) fairness: Lessons for machine learning.
\newblock In \emph{Proceedings of the conference on fairness, accountability, and transparency}, 49--58.

\bibitem[{Jay H.~Hardy et~al.(2022)Jay H.~Hardy, Tey, Cyrus-Lai, Martell, Olstad, and Uhlmann}]{doi:10.1177/0149206320982654}
Jay H.~Hardy, I.; Tey, K.~S.; Cyrus-Lai, W.; Martell, R.~F.; Olstad, A.; and Uhlmann, E.~L. 2022.
\newblock Bias in Context: Small Biases in Hiring Evaluations Have Big Consequences.
\newblock \emph{Journal of Management}, 48(3): 657--692.

\bibitem[{Jiang and Nachum(2020)}]{jiang2020identifying}
Jiang, H.; and Nachum, O. 2020.
\newblock Identifying and correcting label bias in machine learning.
\newblock In \emph{International conference on artificial intelligence and statistics}, 702--712. PMLR.

\bibitem[{Jones et~al.(2017)Jones, Arena, Nittrouer, Alonso, and Lindsey}]{Jones_Arena_Nittrouer_Alonso_Lindsey_2017}
Jones, K.~P.; Arena, D.~F.; Nittrouer, C.~L.; Alonso, N.~M.; and Lindsey, A.~P. 2017.
\newblock Subtle Discrimination in the Workplace: A Vicious Cycle.
\newblock \emph{Industrial and Organizational Psychology}, 10(1): 51–76.

\bibitem[{Kleinberg, Mullainathan, and Raghavan(2016)}]{kleinberg2016inherent}
Kleinberg, J.; Mullainathan, S.; and Raghavan, M. 2016.
\newblock Inherent trade-offs in the fair determination of risk scores.
\newblock \emph{arXiv preprint arXiv:1609.05807}.

\bibitem[{Lahey(2008)}]{Lahey2008}
Lahey, J.~N. 2008.
\newblock Age, {Women}, and {Hiring}: {An} {Experimental} {Study}.
\newblock \emph{Journal of Human Resources}, 43(1): 30--56.

\bibitem[{Langer et~al.(2020)Langer, König, Sanchez, and Samadi}]{automatedinterviews}
Langer, M.; König, C.; Sanchez, D.; and Samadi, S. 2020.
\newblock Highly automated interviews: applicant reactions and the organizational context.
\newblock \emph{Journal of Managerial Psychology}.

\bibitem[{Langer, König, and Papathanasiou(2019)}]{https://doi.org/10.1111/ijsa.12246}
Langer, M.; König, C.~J.; and Papathanasiou, M. 2019.
\newblock Highly automated job interviews: Acceptance under the influence of stakes.
\newblock \emph{International Journal of Selection and Assessment}, 27(3): 217--234.

\bibitem[{Li, Goel, and Ash(2022{\natexlab{a}})}]{10.1145/3514094.3534147}
Li, N.; Goel, N.; and Ash, E. 2022{\natexlab{a}}.
\newblock Data-Centric Factors in Algorithmic Fairness.
\newblock In \emph{Proceedings of the 2022 AAAI/ACM Conference on AI, Ethics, and Society}, AIES '22, 396–410. New York, NY, USA: Association for Computing Machinery.
\newblock ISBN 9781450392471.

\bibitem[{Li, Goel, and Ash(2022{\natexlab{b}})}]{li2022data}
Li, N.; Goel, N.; and Ash, E. 2022{\natexlab{b}}.
\newblock Data-centric factors in algorithmic fairness.
\newblock In \emph{Proceedings of the 2022 AAAI/ACM Conference on AI, Ethics, and Society}, 396--410.

\bibitem[{Lippens, Vermeiren, and Baert(2023)}]{Lippens2023}
Lippens, L.; Vermeiren, S.; and Baert, S. 2023.
\newblock The state of hiring discrimination: {A} meta-analysis of (almost) all recent correspondence experiments.
\newblock \emph{European Economic Review}, 151: 104315.

\bibitem[{Mitchell et~al.(2021)Mitchell, Potash, Barocas, D'Amour, and Lum}]{annurev:/content/journals/10.1146/annurev-statistics-042720-125902}
Mitchell, S.; Potash, E.; Barocas, S.; D'Amour, A.; and Lum, K. 2021.
\newblock Algorithmic fairness: Choices, assumptions, and definitions.
\newblock \emph{Annual review of statistics and its application}, 8(1): 141--163.

\bibitem[{Neumark, Burn, and Button(2019)}]{OlderWorkersFieldExperiment}
Neumark, D.; Burn, I.; and Button, P. 2019.
\newblock Is It Harder for Older Workers to Find Jobs? New and Improved Evidence from a Field Experiment.
\newblock \emph{Journal of Political Economy}, vol. 127, no. 2].

\bibitem[{Newman, Fast, and Harmon(2020)}]{NEWMAN2020149}
Newman, D.~T.; Fast, N.~J.; and Harmon, D.~J. 2020.
\newblock When eliminating bias isn’t fair: Algorithmic reductionism and procedural justice in human resource decisions.
\newblock \emph{Organizational Behavior and Human Decision Processes}, 160: 149--167.

\bibitem[{Pedregosa et~al.(2011)Pedregosa, Varoquaux, Gramfort, Michel, Thirion, Grisel, Blondel, Prettenhofer, Weiss, Dubourg et~al.}]{pedregosa2011scikit}
Pedregosa, F.; Varoquaux, G.; Gramfort, A.; Michel, V.; Thirion, B.; Grisel, O.; Blondel, M.; Prettenhofer, P.; Weiss, R.; Dubourg, V.; et~al. 2011.
\newblock Scikit-learn: Machine learning in Python.
\newblock \emph{Journal of Machine Learning Research}, 12: 2825--2830.

\bibitem[{Pessach and Shmueli(2022)}]{pessach2022review}
Pessach, D.; and Shmueli, E. 2022.
\newblock A review on fairness in machine learning.
\newblock \emph{ACM Computing Surveys (CSUR)}, 55(3): 1--44.

\bibitem[{Rubin(1974)}]{rubin1974estimating}
Rubin, D.~B. 1974.
\newblock Estimating causal effects of treatments in randomized and nonrandomized studies.
\newblock \emph{Journal of educational Psychology}, 66(5): 688.

\bibitem[{Schumann et~al.(2020)Schumann, Foster, Mattei, and Dickerson}]{schumann2020we}
Schumann, C.; Foster, J.; Mattei, N.; and Dickerson, J. 2020.
\newblock We need fairness and explainability in algorithmic hiring.
\newblock In \emph{International conference on autonomous agents and multi-agent systems (AAMAS)}.

\bibitem[{Steiner, Atzmüller, and Su(2016)}]{Steiner2016}
Steiner, P.~M.; Atzmüller, C.; and Su, D. 2016.
\newblock Designing {Valid} and {Reliable} {Vignette} {Experiments} for {Survey} {Research}: {A} {Case} {Study} on the {Fair} {Gender} {Income} {Gap}.
\newblock \emph{Journal of Methods and Measurement in the Social Sciences}, 7(2): 52--94.

\bibitem[{Stredwick(2005)}]{2005116}
Stredwick, J. 2005.
\newblock Chapter 4 - Recruitment.
\newblock In \emph{Introduction to Human Resource Management (Second Edition)}, 116--160. Oxford: Butterworth-Heinemann, second edition edition.
\newblock ISBN 978-0-7506-6534-6.

\bibitem[{Vecchione, Levy, and Barocas(2021)}]{Vecchione_2021}
Vecchione, B.; Levy, K.; and Barocas, S. 2021.
\newblock Algorithmic Auditing and Social Justice: Lessons from the History of Audit Studies.
\newblock In \emph{Equity and Access in Algorithms, Mechanisms, and Optimization}, EAAMO ’21. ACM.

\bibitem[{Verma, Ernst, and Just(2021)}]{DBLP:journals/corr/abs-2102-03054}
Verma, S.; Ernst, M.~D.; and Just, R. 2021.
\newblock Removing biased data to improve fairness and accuracy.
\newblock \emph{CoRR}, abs/2102.03054.

\bibitem[{Wick, Panda, and Tristan(2019)}]{NEURIPS2019_373e4c5d}
Wick, M.; Panda, S.; and Tristan, J.-B. 2019.
\newblock Unlocking Fairness: a Trade-off Revisited.
\newblock In Wallach, H.; Larochelle, H.; Beygelzimer, A.; d\textquotesingle Alch\'{e}-Buc, F.; Fox, E.; and Garnett, R., eds., \emph{Advances in Neural Information Processing Systems}, volume~32. Curran Associates, Inc.

\end{thebibliography}

\end{document}